\documentclass[conference]{IEEEtran}
\IEEEoverridecommandlockouts
\usepackage{cite}
\usepackage{amsmath,amssymb,amsfonts}
\usepackage{algorithmic}
\usepackage{graphicx}
\usepackage{textcomp}
\usepackage{xcolor}
\def\BibTeX{{\rm B\kern-.05em{\sc i\kern-.025em b}\kern-.08em
    T\kern-.1667em\lower.7ex\hbox{E}\kern-.125emX}}
    
\usepackage{tikz}
\newcommand*\circled[1]{\tikz[baseline=(char.base)]{
    \node[shape=circle,draw,inner sep=1pt] (char) {#1};}}
\begin{document}

\title{Improving the Efficiency of Transformers for Resource-Constrained Devices}

\author{\IEEEauthorblockN{Hamid Tabani*\thanks{$*$ Corresponding Author. This paper is accepted as a full paper at 24th Euromicro Conference on Digital System Design (DSD).}, Ajay Balasubramaniam, Shabbir Marzban, Elahe Arani, Bahram Zonooz}
\IEEEauthorblockA{Advanced Research Lab, NavInfo Europe, Eindhoven, The Netherlands\\
\texttt{\{firstname.lastname\}@navinfo.eu, bahram.zonooz@gmail.com}}
}

\maketitle

\begin{abstract}
Transformers provide promising accuracy and have become popular and used in various domains such as natural language processing and computer vision. However, due to their massive number of model parameters, memory and computation requirements, they are not suitable for resource-constrained low-power devices. Even with high-performance and specialized devices, the memory bandwidth can become a performance-limiting bottleneck.
In this paper, we present a performance analysis of state-of-the-art vision transformers on several devices. We propose to reduce the overall memory footprint and memory transfers by clustering the model parameters. We show that by using only 64 clusters to represent model parameters, it is possible to reduce the data transfer from the main memory by more than 4x, achieve up to 22\% speedup and 39\% energy savings on mobile devices with less than 0.1\% accuracy loss.
\end{abstract}


\section{Introduction}
\label{sec:introduction}
Transformers, deep neural network architectures based on self-attention mechanism, were developed to solve the problem of sequence understanding~\cite{vaswani2017attention}. Transformers are widely used for different applications. For instance, they have been used by OpenAI in their language models and by DeepMind in AlphaStar. Although they garnered popularity in other domains such as computer vision~\cite{khan2021transformers}, they come at the cost of having huge number of model parameters.

Therefore, training and inference of transformers are computation-intensive resulting in massive memory usage and energy consumption. BERT (Bidirectional Encoder Representations from Transformers)~\cite{devlin2018bert}, is a transformer-based machine learning model for NLP developed by Google. It has been shown that speeding up the training of BERT by 30\% translates into a savings of over \$85,000 on Amazon Web Service (AWS)~\cite{ivanov2020data}. For the GPT-3 transformer model~\cite{brown2020language} with a training cost of \$12M, this 30\% speedup could save \$3.6M and more than 120 MWh energy~\cite{ivanov2020data}. 
The training process is usually done once using powerful machines while the inference process of the trained model can be deployed many times on resource-constrained devices {\color{black}such as wide range of mobile and wearable devices.} Hence, in this paper, we focus on optimizing transformers for inference.
{\color{black}
Note that it is very challenging to significantly reduce the energy consumption and memory usage of transformers while maintaining the accuracy intact. On the other hand, efficient solutions may only be feasible with adequate hardware support. For instance, to perform the arithmetic operations in 16-bit floating-point (FP16) instead of 32-bit floating-point (FP32), we require the hardware to support FP16 arithmetic. }

Benefits of transformers can be leveraged by deploying them on resource-constrained devices. 
%Deploying transformers on resource-constrained devices can reduce the observed latency and improve privacy for the users over high-cost cloud services. But most of the edge devices are resource-constrained, with memory and compute power limitations.
However, on such devices, memory transactions can become the main bulk of energy consumption. In recent years, application-specific accelerators have become more and more popular due to their efficient performance which make them suitable for low-power mobile and wearable devices. In fact, most of today's devices are integrating tens of accelerators for different tasks~\cite{nextgenAI}. Specialized accelerators and hardware platforms are designed to benefit from application-specific features such as parallelism, quantization, and sparsity due to pruning~\cite{han2016eie}. However, in some cases, the peak performance cannot be achieved due to the memory bandwidth saturation. {\color{black}This is due to the fact that tens of Gigabytes of data needs to be fetched from the main memory and processed in a fraction of a second.} 
This can get worse in heterogeneous system-on-chips (SoC) where multiple computing units such as multicore CPUs, GPUs, and other accelerators access the same memory subsystem simultaneously~\cite{nvidiatx2,Xavier}.

With this understanding about resource-constrained devices, we believe that the efficiency of transformers can be significantly improved. 
Thus, we aim at further improving the performance and energy consumption of transformers for such devices. Our main contributions are as follows: 
\begin{enumerate}
    \item We analyze transformers for inference, identifying key functions which are the most time- and energy-consuming.
    \item Applying clustering schemes on state-of-the-art transformer models for computer vision which reduces the overall size of the parameters and directly impacts the memory bandwidth. 
    \item We show that significant energy and performance improvements can be achieved with negligible impact on the accuracy of the models. We present the existing trade-offs and opportunities to maximize the utilization of specialized accelerators.
\end{enumerate}

The rest of the paper is organized as follows: Section~\ref{sec:background} presents background on transformers. Section~\ref{sec:optimizations} introduces our analysis and approach towards optimization of transformers and the proposed techniques. In Section~\ref{sec:methodology}, we present our methodology and experimental setup. In Section~\ref{sec:results}, we present the results and discuss the existing trade-offs. Section~\ref{sec:relatedwork} summarizes the related work. Finally, Section~\ref{sec:conclusions} concludes the paper.

% Training transformers, specially for domains such as NLP~\cite{devlin2018bert}, is a very computation-intensive task requiring lots of powerful hardware resources. This is due to the huge training data required for such tasks and also the training iterations needed for the model to reach high accuracy.
%All of this is translated into computation power, memory usage, and massive energy consumption. BERT (Bidirectional Encoder Representations from Transformers)~\cite{devlin2018bert} is a transformer-based machine learning model for NLP developed by Google. Speeding up the training of BERT by 30\% translates into a savings of over \$85,000 on Amazon Web Service (AWS). For the GPT-3 transformer model~\cite{brown2020language} with a training cost of \$12M, this 30\% speedup could save \$3.6M and more than 120 MWh energy~\cite{ivanov2020data}.
%It is usually expected that training process is done on high-performance GPU clusters or specialized accelerators such as Google Tensor Processing Unit (TPU)~\cite{jouppi2017datacenter}. Despite this, the inference of large models can be similarly compute-intensive and memory-demanding.

\section{Background}
\label{sec:background}

In this section, we present a brief background on Transformers, their architectural design and features, and their advantages over previous models.

\begin{figure*}[t!]
    \centering
    \includegraphics[width=.8\textwidth]{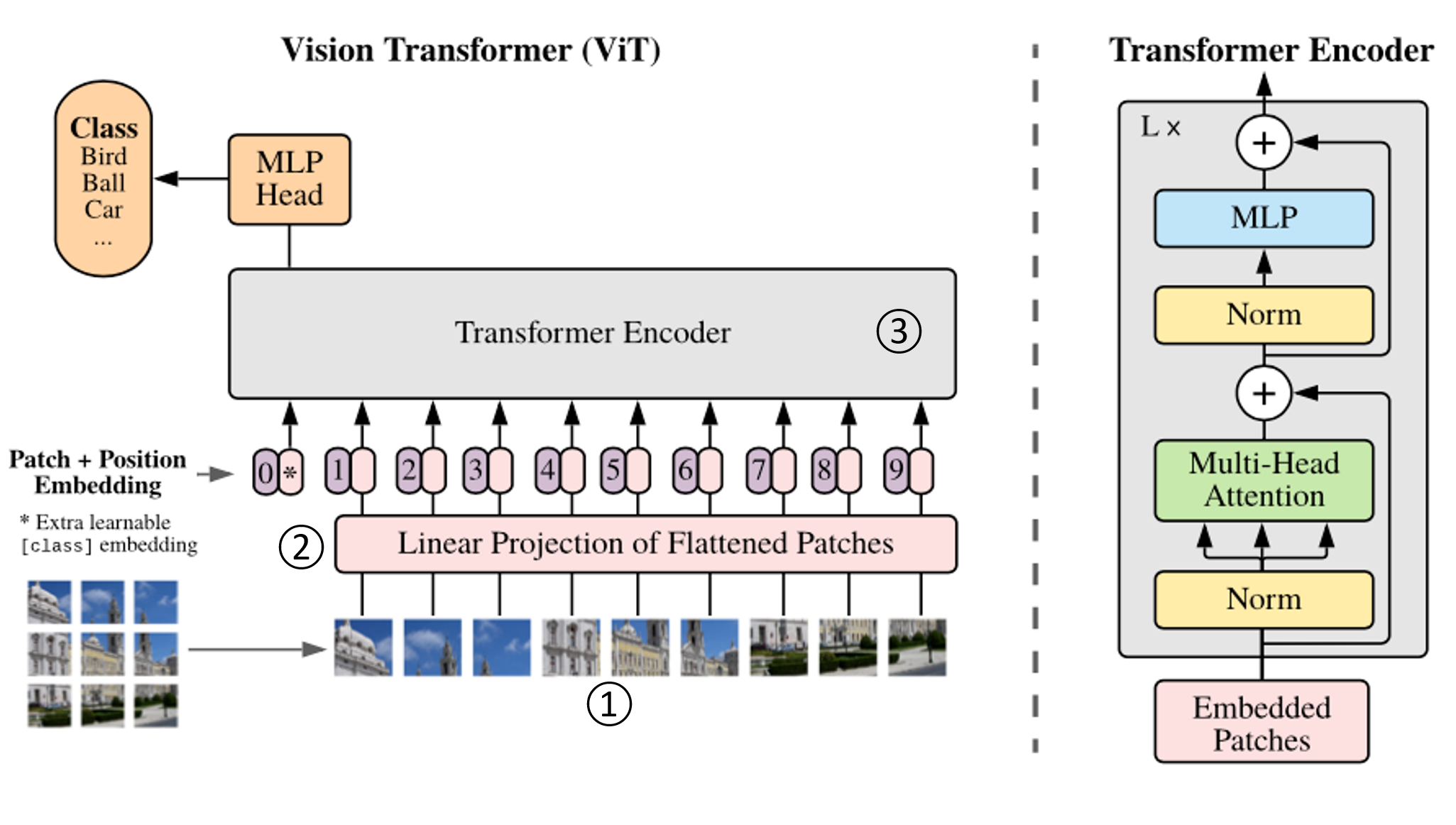}
    \caption{Application of the transformer architecture to image classification~\cite{dosovitskiy2020image}.}
    \label{fig:transformer}
\end{figure*}

%Transformer architecture~\cite{vaswani2017attention} is a neural network model designed for sequence transduction and is transforming an input sequence into an output sequence. Transformers originally developed for machine translation and later used in other domains including Natural Language Processing (NLP) such as language modeling~\cite{wang2019language} and computer vision~\cite{khan2021transformers}.
Before the introduction of Transformers, most state-of-the-art NLP systems relied on gated recurrent neural networks (RNNs)~\cite{rumelhart1985learning}, such as LSTMs~\cite{hochreiter1997long} and gated recurrent units (GRUs)~\cite{chung2014empirical}, with added attention mechanisms. The Transformer architecture is based on these attention technologies but does not use an RNN framework, demonstrating that attention mechanisms alone, without recurrent sequential processing, are powerful enough to achieve RNN-like efficiency. 

Tokens are processed sequentially by gated RNNs, which keep a state vector that contains a representation of the data seen after each token. To process the $n$th token, the model creates a new state that represents the sentence up to token $n-1$ by combining the state representing the sentence up to token $n-1$ with the information from the new token{\color{black}~\cite{chung2014empirical}}. The knowledge from a single token will potentially spread forever.

The implementation of attention mechanisms helped to solve this issue. Attention mechanisms allow a model to look at and draw from the state of the sentence at any point in time. The attention layer has access to all previous states and weighs them according to a learned measure of relevance to the current token, allowing it to provide more precise details about distant related tokens. Translation is an excellent example of the value of focus.
The first word of the French output is almost certainly influenced by the beginning of the English input in an English-to-French translation scheme. In a typical encoder-decoder LSTM model, however, the model is only given the state vector of the last English word to generate the first word of the French output.
Theoretically, this vector can store information about the entire English sentence, providing the model with all required information; however, in practice, this information is frequently lost. When an attention mechanism is added to the model, it will learn to pay attention to the states of early English tokens when generating the beginning of the French production, giving it a much better understanding of what it is translating.

The same issue that affects RNNs in general also affects LSTMs, namely that when sentences are too long, LSTMs perform poorly. The explanation for this is that the likelihood of remembering the meaning of a word far away from the actual word being processed decreases exponentially with time. As a consequence, when sentences are long, the model often forgets the content of positions further down the sequence. Another issue with RNNs and LSTMs is that they are difficult to parallelize for processing sentences since they must be processed word by word. Furthermore, there is no paradigm for long and short-term dependencies. To summarize, LSTMs and RNNs have three issues:
(1) Parallelization is hampered by sequential computation, (2) Long and short-term dependencies are not explicitly modeled, and (3), The “distance” between positions is a straight line.

When attention mechanisms were applied to RNNs, they resulted in significant performance improvements. The Transformer's introduction revealed that attention mechanisms were effective in and of themselves, and that sequential recurrent data processing was not needed to achieve the performance gains of RNNs with attention.
The Transformer, instead of being an RNN, uses an attention system that processes all tokens at the same time and calculates attention weights between them. Transformers can be trained more effectively on larger datasets because they do not rely on sequential processing and lend themselves easily to parallelization.
  
Transformers are multi-layered structures made up of Transformer blocks stacked on top of each other. A multi-head self-attention mechanism, a position-wise feed-forward network, layer normalization modules, and residual connectors distinguish transformer blocks{\color{black}~\cite{tay2020efficient}}. 
Convolutional Neural Networks (CNNs) have inductive biases including translation invariance and a locally limited receptive region, which transformers ignore. Invariance refers to the ability to recognise an entity (i.e. object) in a picture through changes in its presence or location. In computer vision, translation means that each image pixel has been shifted in a certain direction by a specified number. Note that convolution is a linear local operator. Only the neighbor values, as shown by the kernel, are available. The transformer, on the other hand, is permutation invariant by nature and therefore cannot process grid-structured data. Therefore, a spatial non-sequential signal is translated to a series for this purpose. 

Vision transformer is the name of the overall architecture (ViT in short). It first creates patches from a picture and gets the patches as flat as possible \circled{1}. Then, it convert the flattened patches into lower-dimensional linear embeddings \circled{2}. Next, positional embeddings should be added to the sequence and be fed as an input to a standard transformer encoder \circled{3}. Image labels are used to pre-train the model in a fully supervised manner using a huge dataset. Finally, image detection should be fine-tuned on the downstream dataset.
Figure~\ref{fig:transformer} shows how Visual Transformers (ViT) are a straightforward application of the transformer architecture in image classification~\cite{dosovitskiy2020image}.
\section{Performance Analysis and Proposed Optimizations}
\label{sec:optimizations}

In this section, we first present the performance analysis of transformers and then, we present the clustering techniques we applied on transformers for both GPUs and accelerators.

\begin{figure}[t!]
    \centering
    \includegraphics[width=.8\columnwidth]{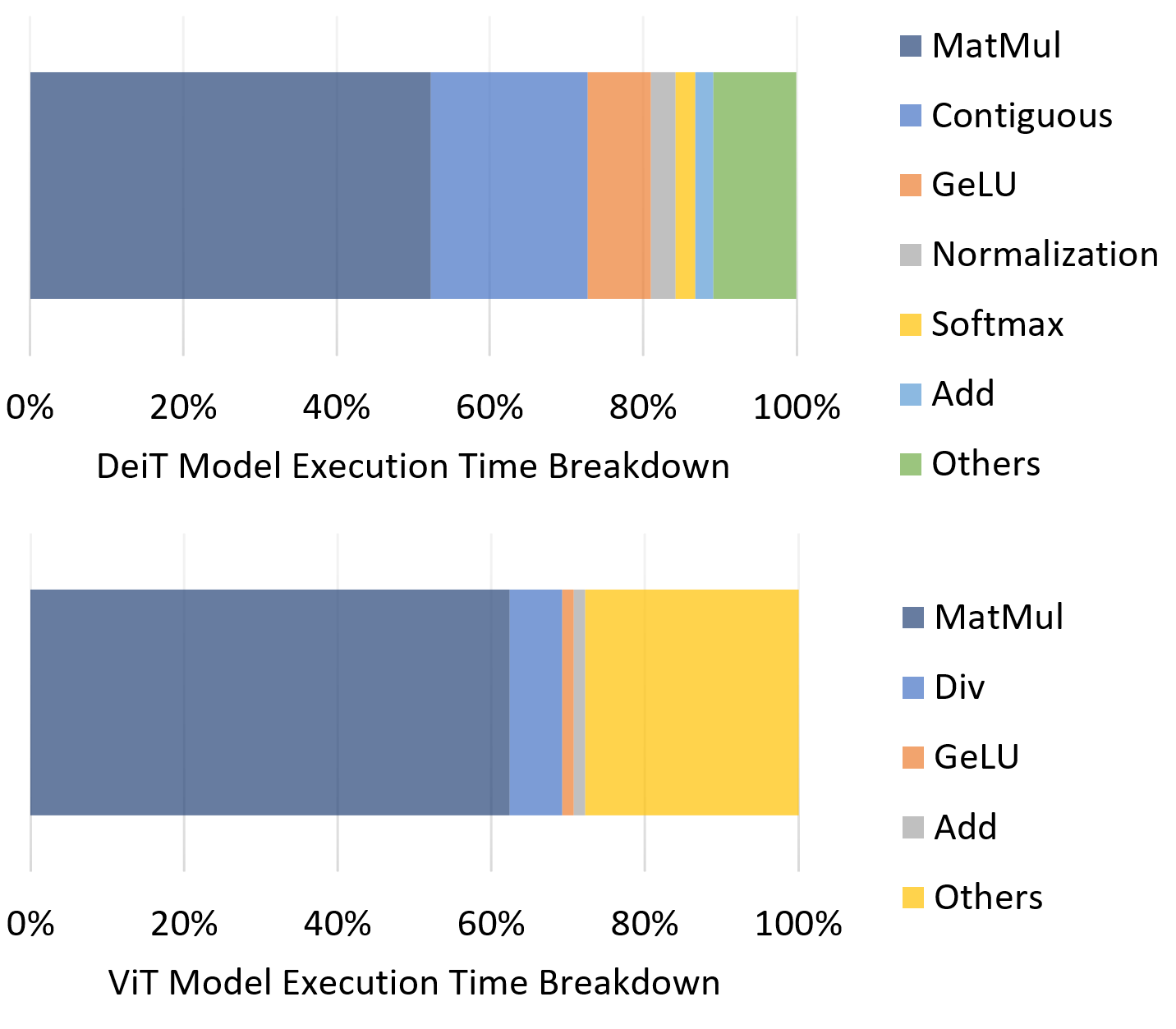}
    \caption{Execution time breakdown of the DeiT and ViT models.}
    \label{fig:exectime-breakdown}
\end{figure}
\begin{figure}[t!]
    \centering
    \includegraphics[width=.755\columnwidth]{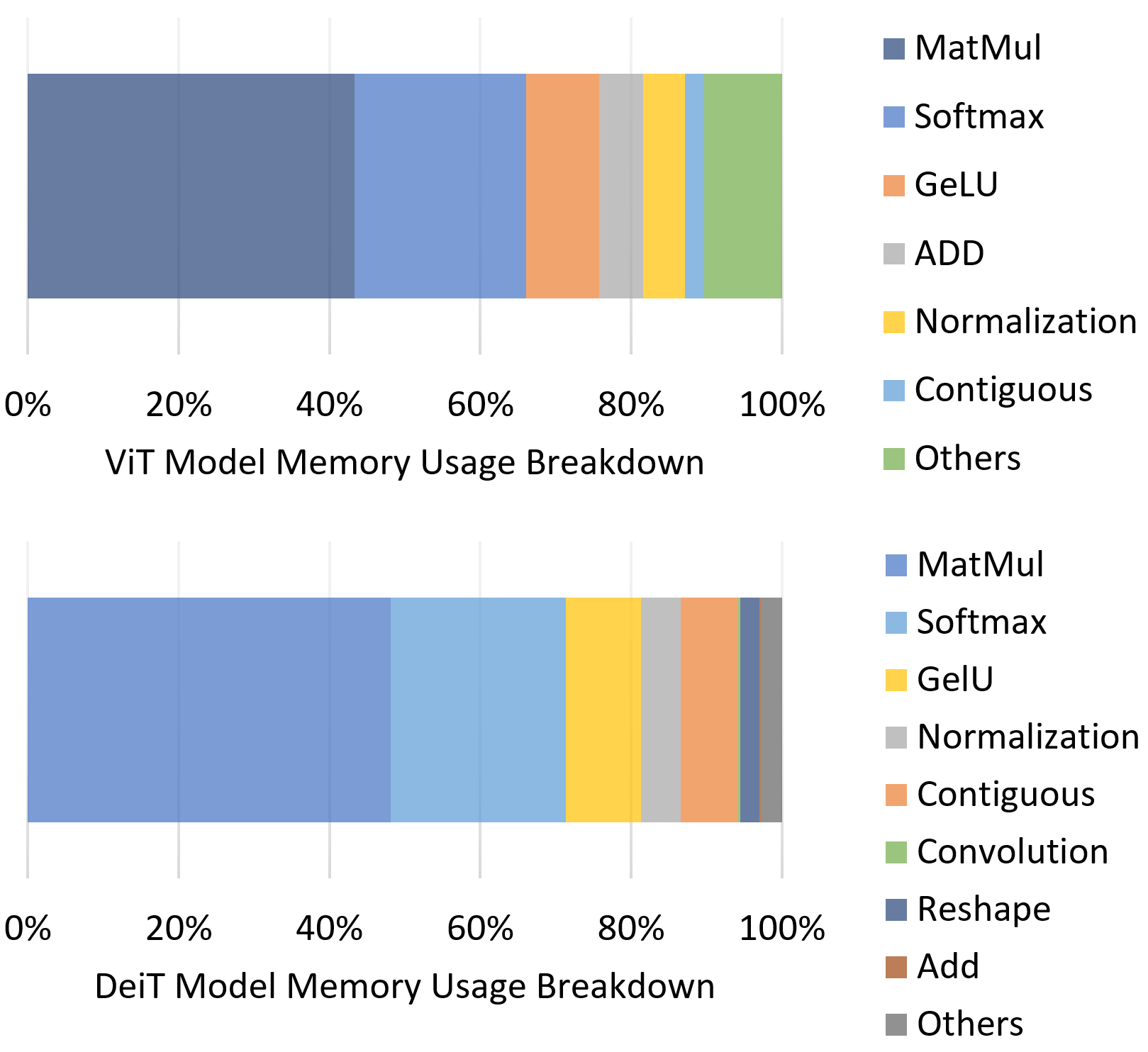}
    \caption{Memory usage breakdown of the DeiT and ViT models.}
    \label{fig:memory-breakdown}
\end{figure}

\subsection{Performance Analysis}
In our experiments, we use two of the latest transformer models for classification:
\begin{itemize}
    \item Classification Transformer (ViT) from Google~\cite{dosovitskiy2020image} which achieves excellent results compared to state-of-the-art convolutional networks. 
    \item DeiT: Data-efficient Image Transformer from Facebook research, a state-of-the-art vision transformer~\cite{touvron2020deit} with 86M parameters.
\end{itemize}

We first profile the two classification models running on the high-end NVIDIA 2080 Ti GPU. The execution time breakdown of different functions of each model running on the GPU is shown in Figure~\ref{fig:exectime-breakdown}. Matrix multiplication is one of the most time-consuming processes during an inference of both models taking more than 50\% of the execution time.

We have analyzed the memory usage for each model as it is shown in Figure~\ref{fig:memory-breakdown}. As the numbers show, in both models, the parameters to perform the matrix multiplication operations are taking more than 40\% of the memory. Next, Softmax and other layers are the most memory-demanding operations as Figure~\ref{fig:memory-breakdown} shows.

\subsection{Clustering}
For both efficiency and energy consumption, reducing the size of the parameters of deep learning models is critical.
The efficiency of inference in transformers is primarily limited by memory transfers. Furthermore, since off-chip memory accesses are one of the main sources of energy consumption in mobile devices, reducing off-chip memory accesses result in significant energy savings{\color{black}~\cite{perrucci2011survey}}.

\begin{figure}[t!]
    \centering
    \includegraphics[width=.95\columnwidth]{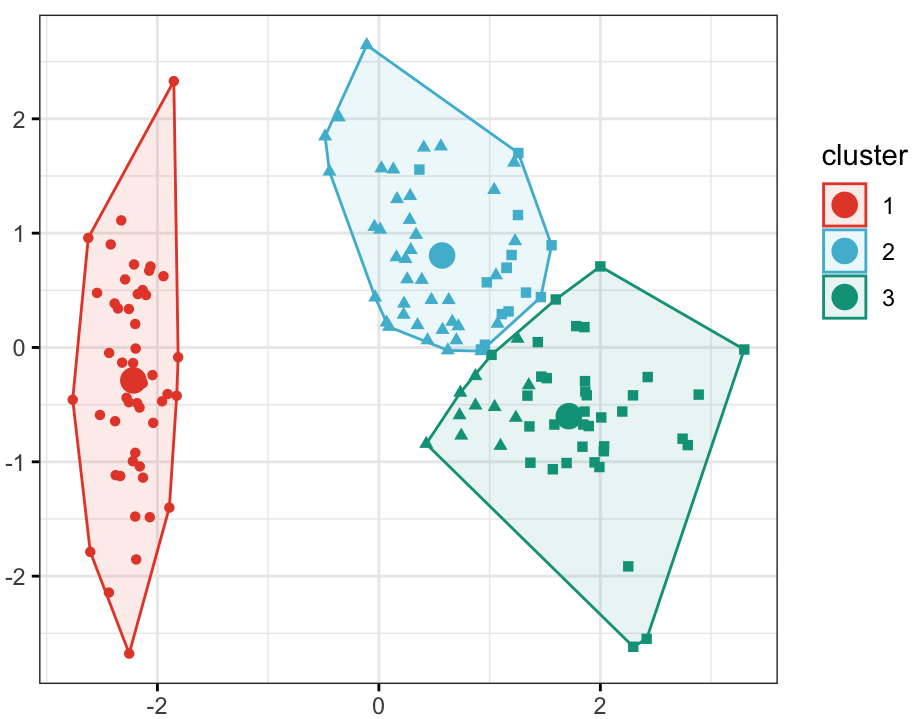}
    \caption{K-means clustering algorithm grouped the data into three clusters each with a centroid.}
    \label{fig:kmeans}
\end{figure}

\begin{figure}[t!]
    \centering
    \includegraphics[width=\columnwidth]{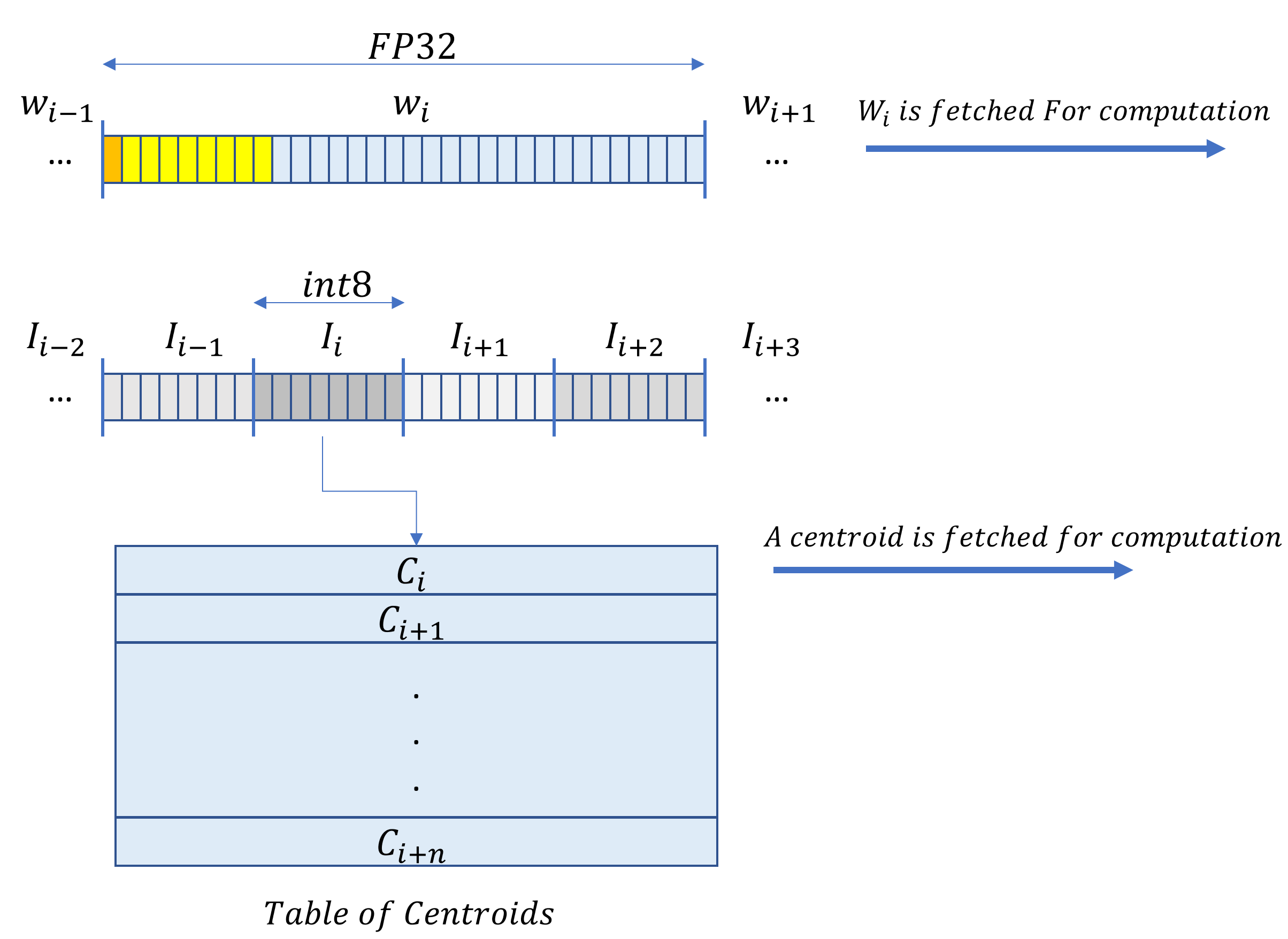}
    \caption{The process of using floating-point 32-bit default parameters vs. using clustered parameters.}
    \label{fig:fp32-clusters}
\end{figure}

Clustering model parameters using K-means~\cite{likas2003global} algorithm is one of the most effective techniques of non-linear quantization for parameter compression. 
Most popular approaches use nonlinear quantization systems, such as K-means, to minimize data size by up to 4 times with negligible change in accuracy~\cite{zafrir2019q8bert,bhandare2019efficient}.
By applying K-means on model parameters, we group the parameters into $k$ clusters each of which has a cluster $centroid$. The centroids are stored in a table referred to as ``table of centroids".
Then, in the step of non-linear quantization, we replace all weights of every cluster by the representative indices from the table of centroids as shown in Figure~\ref{fig:kmeans}. 

Clustering replaces floating-point values in a codebook of centroids with slightly smaller integer indices. When using K-means with 256 clusters, for example, each 32-bit floating-point (FP) value is replaced by an 8-bit index, resulting in a compression ratio of nearly 4x (only 1 KB is required for the storage of a table of 256 32-bit centroids). 
To use the clustered parameters, the corresponding 8-bit index is fetched instead of the 32-bit parameter in the baseline. The index is used to pick the corresponding \textit{centroid} from the very small table of centroids as shown in Figure~\ref{fig:fp32-clusters}.

In this paper, we explore scalar clustering in which every single parameter in the model will be directly represented by an index.
We cluster the model parameters in two ways, as shown in Figure~\ref{fig:clustering-vs-per-layer}:

\begin{figure}[t!]
    \centering
    \includegraphics[width=.75\columnwidth]{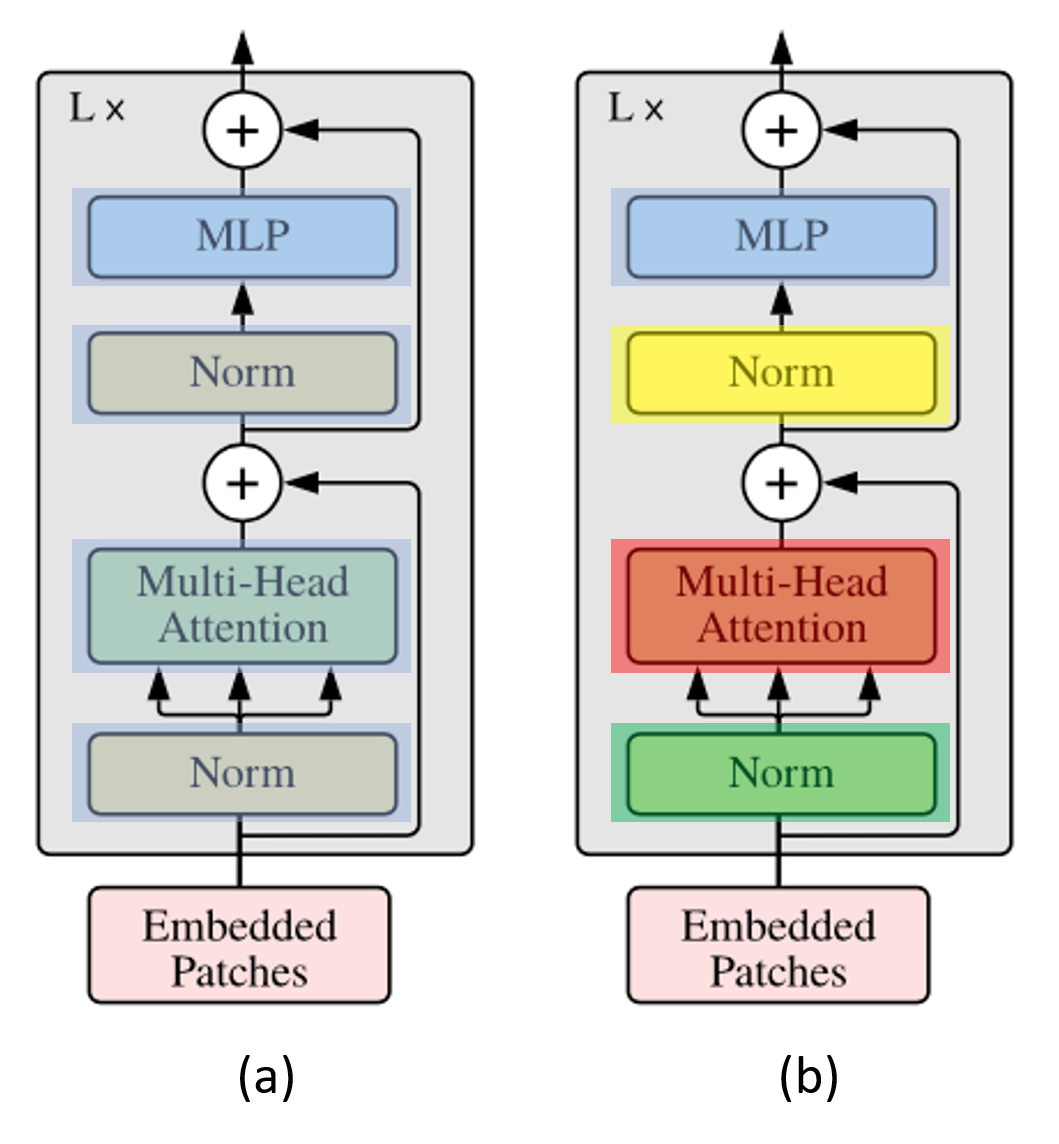}
    \caption{Clustering (a) entire model parameters vs. (b) per-layer clustering.}
    \label{fig:clustering-vs-per-layer}
\end{figure}

\begin{enumerate}
    \item \textbf{Clustering Entire Parameters} in which all the parameters in all the layers are clustered into $c$ number of clusters and a single table of centroids with $c$ entries.
    \item \textbf{Per-Layer Clustering of Parameters} in which the parameters of each individual layer are clustered separately. This means that we will have $c$ number of clusters and a single table of centroids for each individual layer. Assuming $l$ as the number of layers, we will have $l$ separate tables of centroids each with $c$ entries.
\end{enumerate}

{\color{black}As Figure~\ref{fig:clustering-vs-per-layer} (a) shows, parameters in different layers are all sharing the same table of centroids (highlighted in one single color), whereas when performing per-layer clustering, see Figure~\ref{fig:clustering-vs-per-layer} (b), each layer has its own table of centroids (highlighted in various colors).}
Using the clustering techniques, the model size and therefore, memory footprint and memory transfers significantly reduce.
%as shown in Table~\ref{tab:memory-reduction}. 
In the baseline models, parameters are represented using FP32 (single-precision 32-bit floating-point). However, using the clustered parameters, each of the parameters can be represented with only 8 bits to index up to 256 distinct clusters. Although, in theory, less bits are needed to index less number of clusters, e.g., 6 bits for 64 clusters or 5 bits for 32 clusters, however, due to the complexity in alignment and handling data in these formats, in practice, they are rarely used. Therefore, in case of using less number of clusters than 256, the 8-bit index is still used for the sake of simplicity and data alignment in the memory.

%\begin{table}[t!]
%\centering
%\begin{tabular}{  l  c  c  c }
% \hline
% Model & \textbf{Baseline} & \textbf{Clustered} & \textbf{Compression Ratio}  \\ 
% \hline
% DeiT & X MB & Y MB  & Zx  \\
%\hline
%ViT & X MB  & Y MB  & Zx \\
% \hline \\
%\end{tabular}
%caption{Memory usage in the baseline vs. clustered models.}
%label{tab:memory-reduction}
%end{table} 

In the following section, we present the results on various metrics when using clustering techniques on different transformers.
\section{Methodology and Experimental Setup}
\label{sec:methodology}

In this section, we present our methodology and experimental setup employed in this paper.

\subsection{Platform and Hardware Setup}
In order to demonstrate the highest gains of employing clustering schemes when supported in commonly-used platforms, we model three platforms with architectural characteristics similar to the following platforms for our experiments:
\begin{enumerate}
    \item Conf-1: High-end Desktop Configuration. We modeled a desktop system featuring an NVIDIA-like GPU with 4352 CUDA cores and 11 GB GDDR6 memory, similar to a 2080 Ti, an Intel-like processor with 8 cores, and 64 GB of DDR4 Memory.
%\item A hardware accelerator performing the General Matrix Multiplication (GEMM) with the parameters in Table~\ref{tab:accelerator}.
    \item Conf-2: NVIDIA Tegra X2 System-on-Chip (SoC)~\cite{nvidiatx2}. Similarly, we model an NVIDIA TX2-like system featuring a quad-core Arm Cortex-A57 and a dual-core Denver CPU and a 256-core Pascal-based GPU.
    \item Conf-3: NVIDIA AGX Xavier SoC~\cite{Xavier} which features an octa-core Arm-based processor and a 512-core GPU.
\end{enumerate}

Based on our analysis, an efficient hardware modification to support indirect access, which is the key element in implementing clustering schemes, can provide significant benefits in terms of performance improvements and energy consumption.

%\begin{table}[h!]
%\centering
%\begin{tabular}{ | c | l | }
% \hline
% \textbf{Technology / Frequency} & 28 nm / 1500 MHz \\ 
% \hline
% \textbf{Number of CUs / Lane per CU} & 8 / 4 \\  
% \hline
% \textbf{Buffers per CU }& 2 \\
% \hline
% \textbf{Centroid Buffer per CU} & 1 \\
% \hline
%\end{tabular}
%\caption{Hardware accelerator parameters.}
%label{tab:accelerator}
%\end{table}

\subsection{Datasets}
For the evaluations and experimental results, accuracy, and performance analysis in this paper, we use the ImageNet~\cite{ILSVRC15} validation dataset.

\subsection{Evaluation Metrics}
In our evaluations, we consider the following metrics:
\begin{itemize}
    \item \textit{Speedup}. We show the obtained speedup when applying clustering.
    \item \textit{Accuracy}. We compare the relative accuracy change with respect to the baseline model when applying clustering. 
    \item \textit{Memory Usage}. We compare the model size in Megabytes (MB) before and after applying the clustering. Model size reduction is equivalent to compression ratio of the model. With this metric we will demonstrate the usage of the memory bandwidth usage as well as the memory storage requirements to store the model parameters.
    \item \textit{Energy Savings}. We compare the overall estimated energy consumption of the baseline compared to the cluster model. 
\end{itemize}

\begin{figure}[t!]
    \centering
    \includegraphics[width=1\columnwidth]{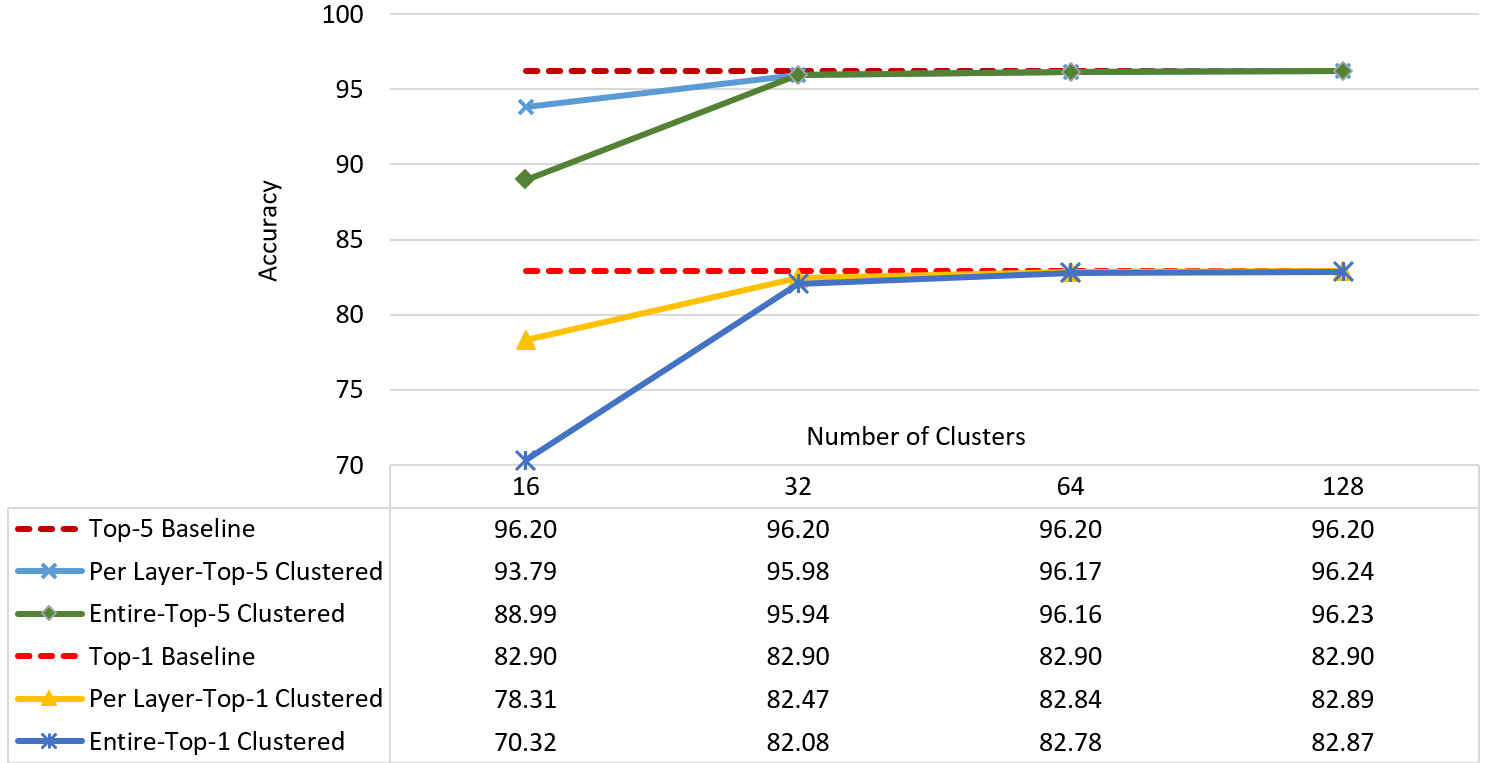}
    \caption{Top-1 and Top-5 accuracy of the DeiT model when using clustering.}
    \label{fig:accuracy-deit}
\end{figure}

\subsection{Implementations}
We use PyTorch~\cite{NEURIPS2019_9015} to run the models and measure the accuracy. We have implemented our own kernels to perform the clustering of the parameters. For each of the GPU platforms, we fine-tune the parameters to gain the best performance.
We use our simulator to measure the performance and timings of the kernels using clustering. To measure the energy consumption accurately for the the platforms similar to NVIDIA TX2 and NVIDIA AGX Xavier platforms, we have accessed the integrated power, current, and voltage rails provided by NVIDIA thermal guide on TX2 and Xavier~\cite{nvidiatx2thermal,nvidiaxavierthermal}. By periodically accessing these rails through specified registers and reading them, we calculate the energy consumption of each unit (e.g., DDR memory, GPU SoC, etc.) for the period of time in which the task is running. We use CACTI 6.5~\cite{thoziyoor2008cacti} to model the energy consumption of the table of centroids.
\section{Experimental Results}
\label{sec:results}

In this section, we first present the results on how clustering techniques affect the accuracy of the different transformer models. Second, we present the performance and energy savings while employing the clustered models. Finally, we discuss the trade-offs between accuracy, memory requirements, performance improvements and energy savings.

\subsection{Accuracy}
\begin{figure}[t!]
    \centering
    \includegraphics[width=1\columnwidth]{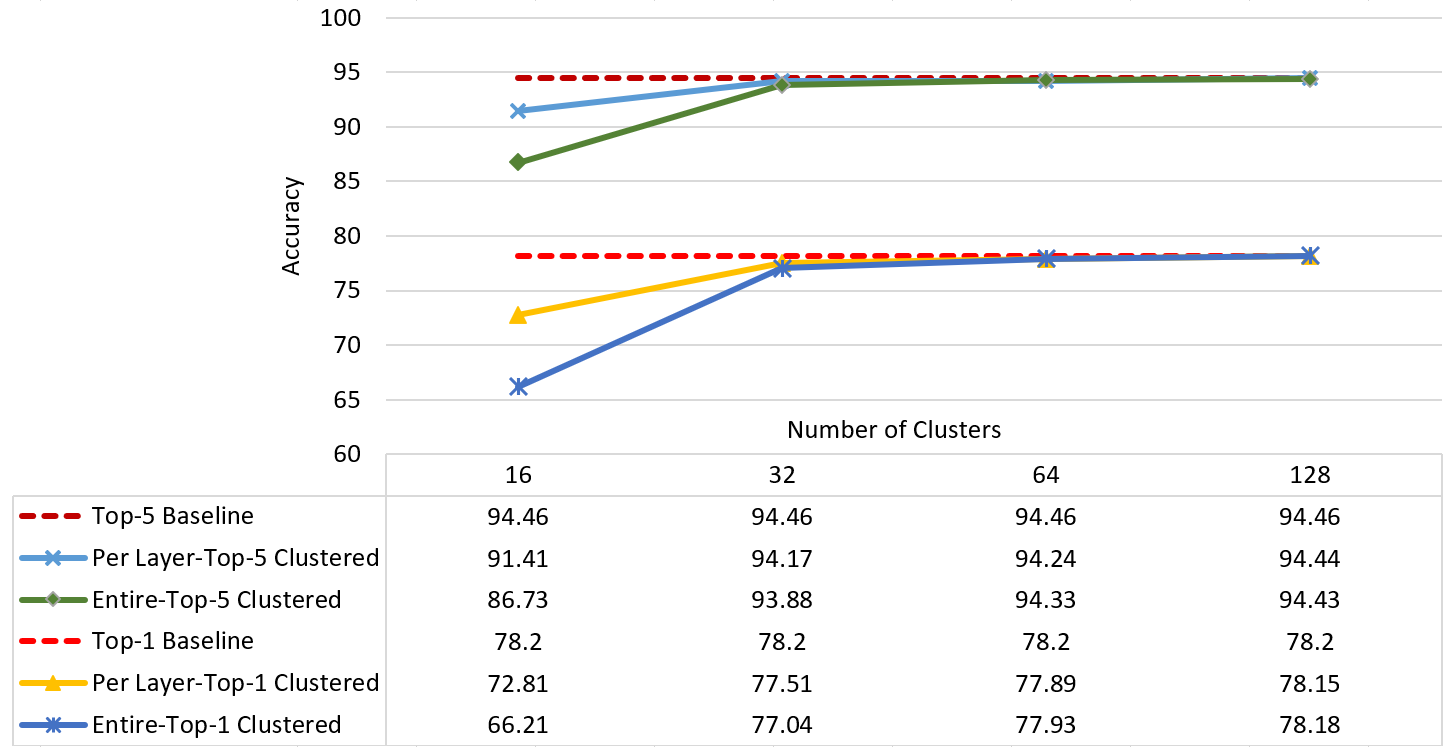}
    \caption{Top-1 and Top-5 accuracy of the ViT model when using clustering.}
    \label{fig:accuracy-vit}
\end{figure}

We applied clustering techniques for the ViT and DeiT models. Figures~\ref{fig:accuracy-deit} and~\ref{fig:accuracy-vit} show the top-1 and top-5 accuracy results for different number of clusters for DeiT and ViT models respectively. Overall, top-1 accuracy results are lower than top-5 results~\cite{touvron2020deit,dosovitskiy2020image}. As Figure~\ref{fig:accuracy-deit} shows, when using less number of clusters, e.g. 16 clusters, the per-layer clustering provides significantly higher accuracy. By increasing the number of clusters the accuracy results reach as good as in the baseline model. For instance, when using 64 clusters for the DeiT model, the top-1 and top-5 accuracy are only 0.1\% and 0.05\% lower than the baseline respectively, which is negligible. By using 128 or more number of clusters, there is zero accuracy loss.

As Figure~\ref{fig:accuracy-vit} shows, we observe similar trend of results, as expected, for the ViT model. We can see that clustering performs the same for this model and by using only 64 clusters for the DeiT model, the top-1 and top-5 accuracy are only {\color{black}0.3\% and 0.2\%} lower than the baseline respectively, which is considered negligible. In these experiments we do not explore more than 128 number of cluster since they provide similar results as the baseline. Similarly, using less than 16 clusters, in most cases, it cannot capture the complexity of the model and results in very high accuracy loss.
Based on our analysis and the related work discussed in Section~\ref{sec:relatedwork}, we conclude that clustering schemes work very well for transformers and they can offer significant benefits.

\subsection{Performance Improvements (Speedup)}
\begin{figure}[t!]
    \centering
    \includegraphics[width=\columnwidth]{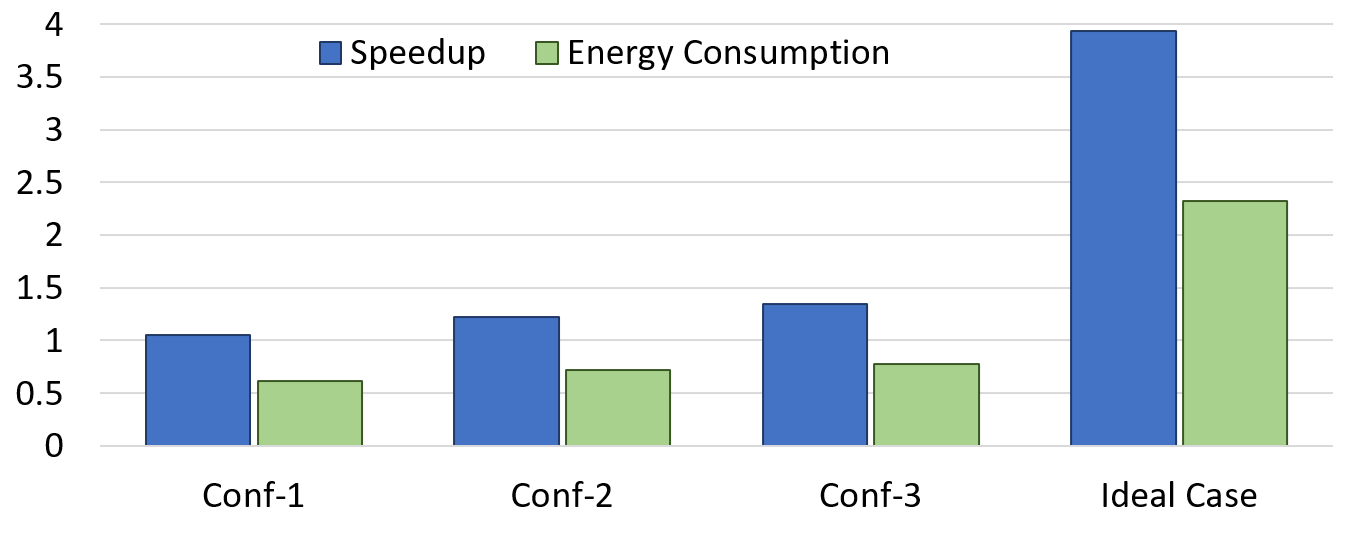}
    \caption{Speedup and energy consumption normalized to the baseline implementation on different modeled platforms.}
    \label{fig:speedup-energy}
\end{figure}

To measure the maximum performance gain when using clustered parameters, we have developed and optimized a specific kernel to operate on clustered data. By fine-tuning and executing our kernel on each of the modeled GPUs, we have observed 5\% to 38\% speedup as shown in Figure~\ref{fig:speedup-energy}. As it shows, despite extra instructions and overhead in the kernel to perform the indirect accesses, as shown in Figure~\ref{fig:fp32-clusters}, the reduced pressure on the memory system, because of clustered parameters, provides significant benefit specially in GPUs with more computing resources such as the GPU in Conf-3.
To demonstrate the advantages of using clustered data, the results are obtained while putting maximum pressure on the memory subsystem. {\color{black}In particular, we have created controlled traffic on the memory subsystem to make limited bandwidth available for our experiments. This is done by concurrently running memory-intensive tasks putting pressure on the memory subsystem. Therefore it leaves less bandwidth for our experiments. As discussed earlier, in modern SoCs, multiple computing units need to access the shared memory.}

\textit{Ideal Case}. To demonstrate the maximum possible performance gains in an ideal system, we have considered a scenario in which the GPU computation power is fully underutilized due to lack of sufficient memory bandwidth. Assuming that the number of computation units is relatively larger than the memory capacity to feed them with data, the GPU can become underutilized. Although in modern GPUs this is a key design factor and it is considered while designing, however, in more powerful and specialized accelerators, this imbalance can become inevitable and therefore, result in performing far from peak capacity.
According to Amdahl's law~\cite{amdahl1967validity}, assuming to have enough computing resources, the reduction in the memory bandwidth and parameters size can increase the computation per memory unit ratio and provide significant performance gains as demonstrated in Figure~\ref{fig:speedup-energy}. We have calculated the maximum speedup possible according to Amdahl's law using an analytical model.

\subsection{Memory Usage}
When applying clustering on the model parameters, the 32-bit parameters are replaced by 8-bit index values. This results in a reduction of 4x in memory usage and memory bandwidth usage.
Note that the table of centroids require a very small memory space relatively. For instance, for 64 clusters, the table of centroids occupies only 256 bytes.

\subsection{Energy Savings}
Figure~\ref{fig:speedup-energy} shows the normalized energy consumption in each of the configuration when using clustering. As it shows in the Conf-1, Conf-2, and Conf-3 the energy consumption reduces by 39\%, 22\%, and 22\% respectively. The highest reduction belongs to the Conf-1 since the memory subsystem takes a considerable portion of the overall energy consumption. The memory subsystem can become the bulk of energy consumption in specialized accelerators~\cite{8091218}. In this case, the energy savings can even provide higher savings overall. 

\textit{Ideal Case}. Similar to the aforementioned discussion regarding the ideal performance, upon achieving the ideal performance, the energy consumption can be drastically reduced as Figure~\ref{fig:speedup-energy} shows. Note that speedup by means of increase in FLOPS usually translates into operating at a higher power. Despite this, due to a shorter execution time considerable static energy is saved in addition to the savings from the reduced memory traffic.

\subsection{Discussion}
The use of clustering model parameters can have numerous objectives and depending on the user's requirements, several factors are considered.

\textit{Accuracy.} Accuracy is the key factor for the user and depending on the task, various thresholds for accuracy loss can be considered in return for performance improvements or energy savings. In other words, depending on user's demands, accuracy loss introduced by using clustering techniques is tolerable. In some cases, achieving the highest accuracy possible is the only target regardless of costs and resource requirements. However, in embedded domain and resource-constrained devices, achieving an acceptable level of performance and resource-usage is always considered. Therefore, such approaches are commonly employed with negligible, and in some cases, with zero accuracy loss.

\textit{Using Generalized Processors vs. Accelerators.}
General-purpose architectures or even GPUs are featuring flexible and programmable design to be used for variety of algorithms and domains. Although it is trivial to implement clustering techniques for CPUs and GPUs, however, it can become costly since the implementations simply translates into more instructions to execute and irregular memory and cache access limiting the performance improvements and in some cases causing slowdown. 
Specialized accelerators, however, can be perfectly designed to obtain highest benefit from approaches such as clustering. Unlike general-purpose processors and GPUs, accelerators will be less flexible by design, however, they can highly perform complicated operations in a more efficient way, if they are designed to do so.

\textit{Applicability in Other Domains.}
In this paper, we focus on the transformers designed and trained for classification tasks, however, as we will discuss in Section~\ref{sec:relatedwork}, the use of clustering schemes and the obtained benefits can be generalized and extended to other domains such as segmentation~\cite{zheng2020rethinking}, Language Modeling~\cite{wang2019language}, other tasks in NLP, and any other domain that transformers are employed. Therefore, similar analysis, discussion, trade-offs, and conclusions can be drawn for those models. It is also proven that the accuracy and the impact of clustering techniques on other domains which are using other forms of deep learning models is similar to our study in this paper. 
%We have presented some of the most relevant related work in Section~\ref{sec:relatedwork}.
\section{Related Work}
\label{sec:relatedwork}

\textbf{Transformers.} Ivanov et al.~\cite{ivanov2020data} state that data movement is the key bottleneck for training of transformers. The massive data requirements during the training, this process becomes memory-bound.
Authors present a recipe for globally optimizing data movement in transformers. Their approach is able to reduce the data movement by more than 20\% while achieving 30\% speedup on BERT~\cite{devlin2018bert}. Although authors claim that their approach is beneficial for other forms of transformers, it is not clear to which extend those models get benefit. 
{\color{black}
In this work, we perform a totally different approach which is orthogonal to the approach presented in~\cite{ivanov2020data}. Furthermore, our approach provides manifold performance and energy improvements while drastically reducing the memory bandwidth requirements.}

Polino et al.~\cite{polino2018model} studied the effect of quantization which is mainly focusing on training. They proposed a technique called quantized distillation and leverage distillation during the training process, and a second technique, differentiable quantization, to optimizes the location of quantization points through stochastic gradient descent, to better fit the behavior of the model. Other works also performed similar studies and employed quantization during the training process or as a post-training step~\cite{fan2020training,wu2020integer,bai2020binarybert}. {\color{black} In this paper, we target the trained models and we proposed a different clustering scheme. Also, we focus on resource-constrained devices and provide extensive performance and energy analysis.} 

Shen et al.~\cite{shen2020q} studied different forms of quantization on deep bidirectional transformers BERT~\cite{devlin2018bert}. They performed an extensive analysis of fine-tuned BERT models using second order Hessian information in order to propose a novel method for quantizing BERT models to ultra low precision.
In~\cite{bhandare2019efficient}, authors quantized a trained Transformer machine language translation model to lower precision 8-bit integers. They also reported performance analysis for Intel MKL Library. In~\cite{boo2020fixed}, authors propose to use fixed-point arithmetic. The fixed-point optimization steps consist of quantization sensitivity analysis, hardware conscious word-length assignment, quantization and retraining, and post-training for improved generalization.

\textbf{Other Models.} 
Han et al.~\cite{han2015deep} performed model pruning and post-training clustering, and Huffman coding to compress multiple computer vision models. They further proposed an specialized accelerator~\cite{han2016eie} to better explore the sparsity and the clustered parameters and achieved four orders of magnitude energy reduction over the baseline models running on a CPU.
Tabani et al.~\cite{8091218,tabani2018low} designed an accelerator for acoustic scoring process in hybrid large-vocabulary speech recognition systems. They show that despite orders of magnitude performance and energy consumption improvements using their design, clustering the parameters provides additional manifold speedup and energy savings. {\color{black} These works target different models such as CNNs or other domains such as speech recognition which are very different than transformers.} 

Similar to previous deep learning models, quantization schemes are well studied for transformers, including for both training and post-training processes, however, we believe that studies remained focused on commonly-used schemes such as vector and scalar quantization rather than techniques such as clustering of the data. Furthermore, the main focus of the previous work is on the accuracy and reasoning behind accuracy change when applying those techniques while in this work, we devote an important part of the paper to performance-related aspect. We also consider variety of frameworks and heterogeneous platforms which, to our knowledge, has not been explored earlier.
\section{Conclusions}
\label{sec:conclusions}

%In recent years, we are witnessing monumental growth in the use-cases and capability of transformers in variety of domains such as NLP and computer vision as they provide promising precision. They are not, however, ideal for resource-constrained low-power systems due to their large memory and computing requirements.
In this paper, we first performed an extensive analysis of state-of-the-art vision transformers. We showed that applying clustering techniques on these models can provide significant speedup and energy savings on various platforms while having negligible drop in the overall accuracy. We presented the existing trade-offs and challenges towards improving the performance and energy savings and how it can be generalized to transformers trained for other domains.
Our results on representative platforms show that significant energy savings is achievable while highest speedups can be reached in specialized accelerators.

\newpage
\bibliographystyle{IEEEtran}
\bibliography{bibliography}

\end{document}